%% file: main.tex
\pdfoutput=1
\documentclass{article}
\usepackage{iclr2027_conference,times}
\input{math_commands.tex}

\usepackage{hyperref}
\usepackage{url}
\usepackage{graphicx}
\usepackage{booktabs}
\usepackage{multirow}
\usepackage{amsmath,amssymb}
\usepackage{xspace}
\usepackage{minitoc}
\usepackage{xcolor}
\usepackage{caption}

\usepackage[capitalize]{cleveref}
\crefname{section}{Sec.}{Secs.}
\Crefname{section}{Section}{Sections}
\crefname{table}{Tab.}{Tabs.}
\Crefname{table}{Table}{Tables}

\newcommand{\methodname}{PAMI\xspace}
\newcommand{\vaeName}{PamiVAE\xspace}
\newcommand{\generatorName}{PamiGen\xspace}
\newcommand{\refineName}{PamiRefiner\xspace}

\title{\textbf{\methodname}: \textbf{\underline{P}}art \textbf{\underline{A}}nchored \textbf{\underline{M}}otion for Text to Human-Object \textbf{\underline{I}}nteraction Generation}

\author{
\begin{tabular}{@{}l@{\hspace{1.4em}}l@{\hspace{1.4em}}l@{\hspace{1.4em}}l@{\hspace{1.4em}}l@{}}
Chuqiao Li\textsuperscript{1} & Xianghui Xie\textsuperscript{1,2} & Yong Cao\textsuperscript{1} & Andreas Geiger\textsuperscript{1} & Gerard Pons-Moll\textsuperscript{1,2}
\end{tabular}
\vspace{-4mm}
\\ \\
{\small \textsuperscript{1}Tübingen AI Center, University of Tübingen, Germany}\\
{\small \textsuperscript{2}Max Planck Institute for Informatics, Saarland Informatics Campus, Germany}\\
{\small \url{https://coral79.github.io/pami/}}
}

\iclrfinalcopy

\begin{document}
\doparttoc
\faketableofcontents

\maketitle

\begin{figure}[h!]
    \centering
    \includegraphics[width=0.99\textwidth]{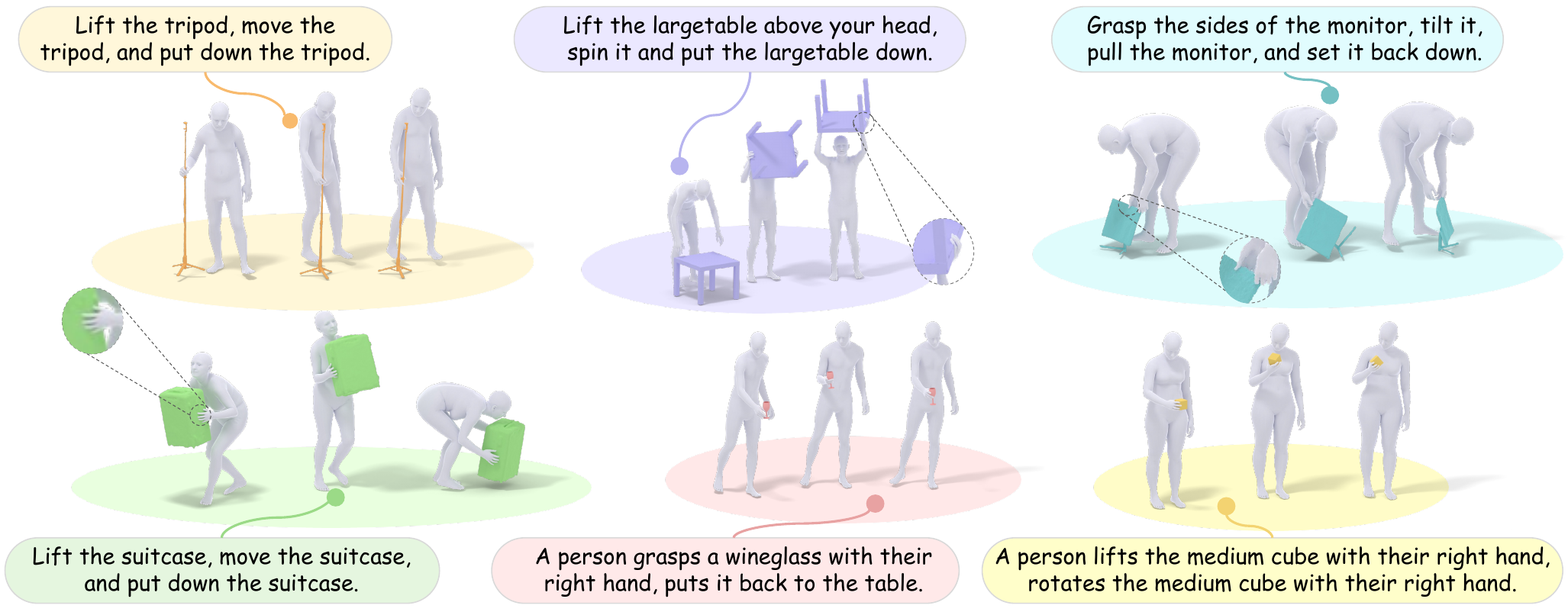} 
    \caption{\textbf{Text-conditioned full-body HOI generation.} Given a text prompt and a
    canonical object mesh, \methodname generates coordinated human and object motion across diverse
    one-handed, two-handed, and full-body interactions. Insets highlight fine-grained contact.}
    \label{fig:teaser}
\end{figure}

\begin{abstract}
\input{sections/abstract}
\end{abstract}

\input{sections/intro}
\input{sections/related}
\input{sections/method}
\input{sections/experiments}
\input{sections/conclusion}

\subsubsection*{Acknowledgments}
Special thanks to the RVH and AVG members for their help and discussions.
Prof. Gerard Pons-Moll and Prof. Andreas Geiger are members of the Machine Learning Cluster of
Excellence, EXC number 2064/1 -- Project number 390727645. Gerard Pons-Moll is endowed by the Carl
Zeiss Foundation. Andreas Geiger was supported by the ERC Starting Grant LEGO-3D (850533).
Yong Cao was supported by a VolkswagenStiftung Momentum grant.

\subsubsection*{AI use statement}
We used large language models (LLMs) to assist with code implementation and to polish the
writing of this paper. All AI-assisted content was reviewed and verified by the authors, who
take full responsibility for the content of this work.

\subsubsection*{Reproducibility statement}
We will publicly release our full code, including training, inference, and evaluation scripts,
to ensure that all results in this paper are fully reproducible.

\bibliography{iclr2027_conference,refs}
\bibliographystyle{iclr2027_conference}

\appendix
\input{sections/appendix}
\end{document}

%% file: math_commands.tex
\usepackage{amsmath,amsfonts,bm}

\def\eqref#1{equation~\ref{#1}}

\def\1{\bm{1}}

\DeclareMathAlphabet{\mathsfit}{\encodingdefault}{\sfdefault}{m}{sl}
\SetMathAlphabet{\mathsfit}{bold}{\encodingdefault}{\sfdefault}{bx}{n}



%% file: sections/abstract.tex
Text-conditioned full-body human--object interaction (HOI) generation requires synthesizing human motion and object trajectories that match the input text while remaining precisely coordinated over time. Most methods represent the human and object as separate trajectories and predict the global human-object couplings. 
Learning this complex, dynamically changing relationship implicitly, however, often yields object drift, missed contact, and penetration.
We introduce \methodname, a Part-Anchored Motion framework for Interaction generation. Inspired by the classic Hough Transform, our key idea is to localize object motion by letting body-part anchors vote for it: we express object motion relative to multiple body-part anchors and use \vaeName to learn an interaction latent space, decoding frame-wise weights that aggregate these part-specific votes. Building on this representation, \methodname generates interactions in a coarse-to-fine hierarchy. \generatorName first generates a coarse human--object interaction from text in this structured latent space, and \refineName then recursively resolves fine-grained contact geometry using a hybrid surface-sensing representation, combining long-range probes that capture overall body-part influence with short-range sensors that resolve detailed contacts near the object surface. Experiments on InterAct show that \methodname generates more faithful interactions and more accurate human-relative object motion than previous methods, achieving 14.5\% higher contact recall than the previous state of the art. Extensive ablations validate the contributions of both the part-anchored voting representation and hybrid surface-sensing refinement.

%% file: sections/intro.tex
\section{Introduction}

Humans constantly interact with objects, making the ability to generate human--object interaction (HOI) a cornerstone of physical intelligence and embodied AI, with direct implications for robotics, animation, and AR/VR. Driving this generation from natural language is especially valuable, as text offers an intuitive and scalable interface for specifying the rich space of interactions without motion capture or manual authoring. Text-conditioned full-body HOI generation is however non-trivial: it must produce a human motion and an object trajectory that match the text and stay coordinated over time. This makes HOI more than the independent generation of two plausible motions, as small relative errors can produce a floating grasp, a drifting object, missed contact, or penetration.

Most full-body text-to-HOI methods diffuse explicit per-frame human and object states \cite{xu2023interdiff, xu2025interact, wang2026unleashing}.  In
text-to-motion, learned temporal representations have shown that motion can instead be generated in
a compact latent space~\citep{chen2023executing,zhang2023t2m,jiang2023motiongpt,li2024unimotion}. Recent work HOIMask~\cite{ji2026hoimask} is the first to use vector quantized latents for full body interaction. 
However, it represents interaction as two separate trajectories, which cannot capture the complex human-object relations in latent space. Similarly, most other methods~\cite{wang2026unleashing} also use the separate human-object trajectories as representation and rely on feature exchange~\cite{wang2026unleashing}, contact prediction~\cite{diller2023cg}, or guidance based correction~\cite{peng2023hoi} to reason about the interactions. Some methods explore predicting contact distances~\cite{xu2025interact} or weighted combination of multiple predictions based on distances~\cite{cha2024text2hoi, xu2025interact}. But such feature space based correlation reasoning or hard coded weights is suboptimal as the network needs to reason about the complete object geometry under all dynamic pose transformations and even from body regions that are not affecting the object, making it hard to learn the essential correlations efficiently. In this paper, we ask \textit{what representation is suitable for generating interaction motion in latent space while respecting the intricate human-object interaction and contacts?}

To this end, we introduce \textbf{\methodname}: \textbf{\underline{P}}art \textbf{\underline{A}}nchored \textbf{\underline{M}}otion voting for \textbf{\underline{I}}nteraction motion generation in latent space. We take inspiration from the classic Hough Transform~\citep{ballard1981generalizing} in computer vision, where a global structure that is hard to detect directly is instead localized by letting many local elements each vote for it and taking their consensus, an aggregation that is robust to noise and to irrelevant votes. We cast object motion in the same way: the object pose is the global structure, and the body-part anchors are the local voters. We express object translation relative to each body-part anchor in its local frame, so every anchor votes for where the object should be, and let the decoder predict how strongly each anchor votes per frame. Combining these votes with learned weights forms a soft, differentiable Hough aggregation that naturally down-weights parts not involved in the interaction. We embed this representation with \vaeName, a variational autoencoder (VAE), into a compact interaction latent that preserves subtle geometric relations yet is easier to generate. 

Human--object interaction motion is highly dynamic and complex especially from only text as input. We therefore adopt a hierarchical coarse-to-fine design, first recovering an approximate interaction and then resolving its fine-grained contact geometry. In the coarse stage, we train \generatorName, a text-conditioned generator that produces interaction latents decoded into a human--object motion that captures the text, the overall movement, and the anchor-voted human--object relation. In the fine stage, \refineName recursively refines this motion through hybrid local geometry sensing: long-range probes capture overall body-part influence, while short-range sensors capture detailed contact in the immediate vicinity of the surface. This decomposition lets latent voting establish the coarse interaction while surface sensing recovers the fine, contact-rich geometry.

Experiments on InterAct~\cite{xu2025interact} dataset show that our method generates more faithful interaction motions with more accurate object motions relative to the human than prior arts. Notably, \methodname achieves 14.5\% higher contact recall than previous state-of-the-art method. Extensive ablations also validate that our part-anchored representation is important to learn a meaningful latent space for HOI generation and that our hybrid surface sensing and refinement improves geometric correctness of the generated motions. 
In summary, our main contributions are:
\begin{itemize}
\item We present \methodname, to our knowledge, the \textbf{first part-based latent generative framework} for text-conditioned full-body HOI motion generation. Our generation is hierarchical from coarse to fine and achieves the best performance on the benchmark and generates high-quality interaction motion from text. Our code and model will be publicly released.

\item We introduce a \textbf{body-part based object voting representation} that anchors object motion to body parts and lets anchors vote for object movement via learned weights. This allows efficient generation in latent space while preserving subtle human-object relations, yielding semantically well aligned coarse interaction motion.  

\item We design a \textbf{hybrid local surface sensing module} to train a RefineNet that reasons fine human-object geometry and contacts to improve the coarse interaction motion from latent generation. Trained with mixed data from generation and ground truth, our RefineNet improves not only our latent generation results but also baseline generations zero-shot.

\end{itemize}

%% file: sections/related.tex
\section{Related Work}
\textbf{Text-to-HOI and latent motion generation.}
Text-conditioned interaction generation has been studied for hand--object motion~\citep{cha2024text2hoi,christen2024diffh2o,song2026jointhoi} and for full-body human--object motion. Early full-body methods condition generation on intent or object trajectories~\citep{ghosh2022imos,li2023object,li2023controllable}; more recent diffusion and related generative models generate interactions from text and incorporate contact, object geometry, or human--object relation features~\citep{xu2023interdiff,peng2023hoi,diller2023cg,wu2024thor,song2024hoianimator,xu2024interdreamer,zeng2025chainhoi,xue2025guiding,geng2025auto,li2025genhoi,jung2026decoupled,cai2026vihoi,wang2026mami,wu2026hoi,di2026pamotion,zou2026infbagel}. HIMO~\citep{lv2024himo} handles multiple objects, InterAct~\citep{xu2025interact} introduces a marker-based representation and a large-scale benchmark, and LIGHT~\citep{wang2026unleashing} denoises body, hand, and object streams at different paces. Most full-body methods operate directly on per-frame motion states, coupling the human and object through feature exchange, relation or contact prediction, or sampling guidance. LatentHOI~\citep{Muchen_LatentHOI} uses latent diffusion for hand--object motion, but does not model full-body interaction. In text-to-motion, learned tokenizers and temporal autoencoders have made generation in compact latent spaces common~\citep{chen2023executing,zhang2023t2m,jiang2023motiongpt,li2024unimotion,nazarenus2026actionplan}. Whole-body and part-based models further separate the dynamics of different body regions~\citep{humantomato,li2026frankenmotion}. We design a novel part-anchored voting representation and embed it into a compact latent space for faithful interaction generation.

\textbf{Local and relative object representations.}
Relative object representations encode human--object coordination directly~\citep{li2023object,zeng2025chainhoi}. Local modeling also appears in STAR~\citep{STAR:2020}, which restricts pose-dependent body deformation to sparse joint neighborhoods rather than coupling every joint to the full surface. In HOI generation, InterAct~\citep{xu2025interact} uses vectors from body markers to the nearest object surface and weights relative object-motion guidance by marker--object distance. CG-HOI~\citep{diller2023cg} predicts object transformations associated with body-surface markers and combines them using contact-distance weights. In both cases, distance determines which human locations influence the object. Dex2HOI~\citep{pratikaki2026dex2hoi} instead predicts mixture weights for left-wrist-relative, right-wrist-relative, and free object trajectories. Our representation extends this learned composition from wrists to full-body and embed them in latent space.

\textbf{Guidance and refinement for HOI.}
Physical errors in generated interactions can be addressed during training, sampling, or after generation. Contact-aware losses and auxiliary contact or affordance prediction supervise the generator~\citep{li2023controllable,ma2024contact,peng2023hoi,diller2023cg,cha2024text2hoi,song2026jointhoi,di2026pamotion}, while guidance and relation intervention modify diffusion sampling~\citep{karunratanakul2023gmd,wu2024thor,xue2025guiding,wang2026unleashing,li2025hoi,li2026conta,ron2026hoidini}. InterDiff~\citep{xu2023interdiff} performs kinematics-informed iterative correction, and OOD-HOI~\citep{zhang2024oodhoi} uses predicted contact regions during diffusion inference. Other work applies optimization~\cite{xie2022chore, xie2023visibility,lee2026mochi}, learned~\cite{zhou2022toch,taheri2024grip, xie2024template,wang2026mami} or physics-based correction~\citep{zhang2024manidext,xu2024interdreamer,xu2025intermimic,xu2026interprior,lin2025simgenhoi}. GEARS~\citep{zhou2024gears} and GRAFT~\citep{ym2026graft} use local geometry for hand--object synthesis and human--scene fitting, respectively. We follow the widely used hierachical design and use latent generation for coarse and local geometry sensing for fine interaction generation.

%% file: sections/method.tex
\section{Method}

\begin{figure*}[t]
    \centering
    \includegraphics[width=0.99\textwidth]{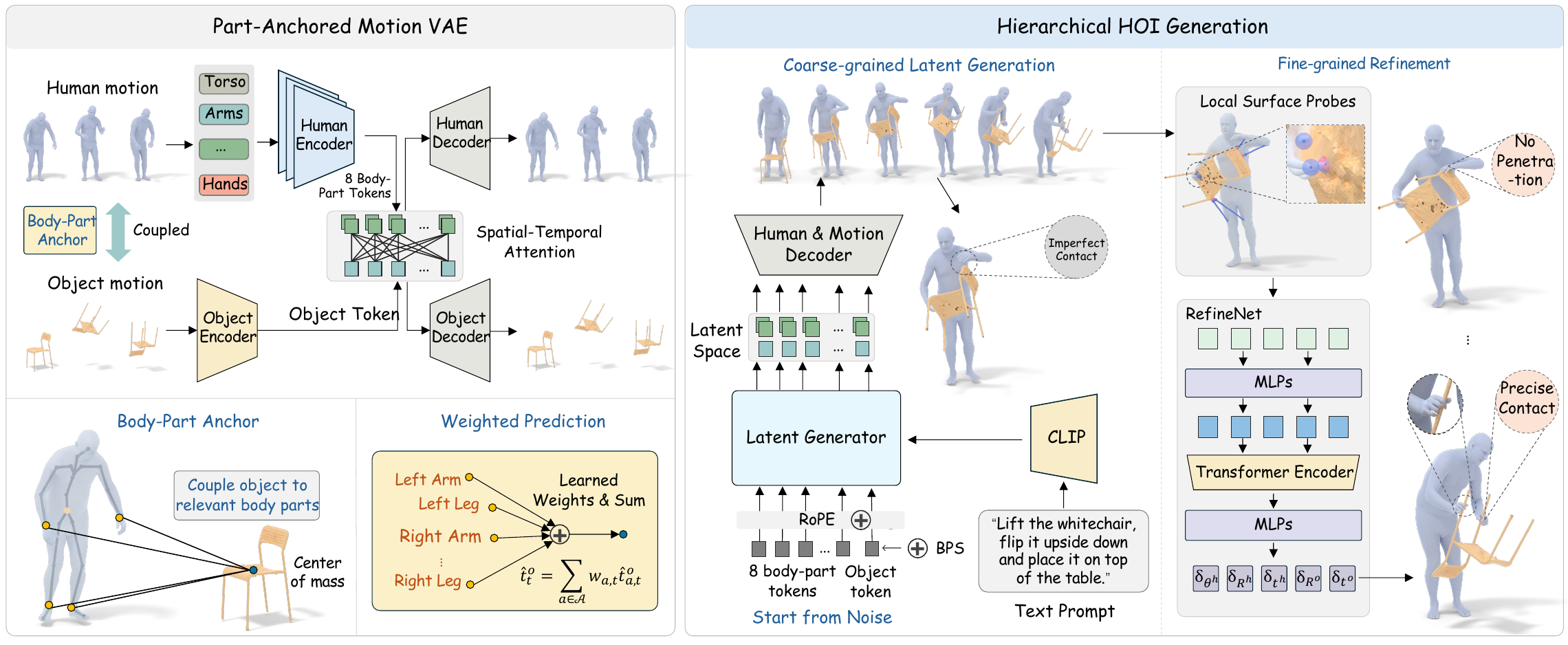} 
    \caption{\textbf{Overview of \methodname.} Body-part anchors vote for the object motion, and \vaeName embeds this part-anchored interaction into a latent space (\cref{sec:structured-latent}). Generation then proceeds coarse-to-fine: \generatorName samples a coarse interaction from text in the latent space (\cref{sec:latent-generation}), which \refineName refines in explicit space using hybrid long- and short-range surface-sensing probes to sharpen contacts (\cref{sec:refinement}).
    }
    \label{fig:method}
\end{figure*}

Given a text prompt $c$, a canonical object mesh $\mathcal{M}$, our goal is to generate a human motion $\mathbf{x}^{h}_{1:T}$ and a rigid-object trajectory
$\mathbf{x}^{o}_{1:T}$ of $T$ frames that form one coordinated interaction. We use SMPL-H~\cite{loper2015smpl,MANO} to represent human motion. 
Since the generation must jointly satisfy the text, the human motion, and their fine-grained coupling, we adopt a coarse-to-fine design, illustrated in \cref{fig:method}. \vaeName first learns a compact interaction latent space built on our part-anchored voting representation (\cref{sec:structured-latent}). A text-conditioned generator \generatorName then produces a coarse interaction in this space (\cref{sec:latent-generation}). Finally, \refineName operates in explicit space and recursively refines this interaction to recover fine-grained contact geometry (\cref{sec:refinement}).

\subsection{Part-Anchored Interaction VAE}
We aim at compressing the Euclidean-space interaction into a compact latent capturing the human-object motions and their coupling. We use a variational autoencoder to encode the two with separate body-part and object encoders then couple them via spatiotemporal attention, and decodes the human motion together with a part-anchored voting representation of object motion, detailed next.

\label{sec:structured-latent}
\label{sec:part-anchor}
\textbf{Part-anchored object motion representation.} Because object motion is most strongly coupled to contacting or supporting body parts, we predict it in the spirit of the Hough Transform: each body part votes for object motion, and per-frame weights aggregate the votes. For detached objects, no body part should drive motion, so we add a free anchor that votes independently of the human.

For each sequence, we define the world coordinate frame as the human root coordinate in the first frame and express all subsequent global human and object motion relative to this frame. We write the rigid-object trajectory to be generated as
$\mathbf{x}^{o}_{t}=(\mathbf{R}^{o}_{t},\mathbf{t}^{o}_{t})$, where
$\mathbf{R}^{o}_{t}\in\mathrm{SO}(3)$ and $\mathbf{t}^{o}_{t}\in\mathbb{R}^{3}$ denote the object
orientation and translation at frame $t$, respectively. Let $\mathcal{A}$ denote a set of $K=6$ body-part anchors, which includes the selected body joints~\cite{li2026frankenmotion} of SMPL model, and one additional free anchor that is the same as the world space. Let $\mathbf{p}^{h}_{a,t}, \mathbf{R}^{h}_{a,t}$ be the local coordinate frame of an anchor $a\in \mathcal{A}$, 
we represent the object translation relative to each anchor as: 
$\boldsymbol{\Delta}^{o}_{a,t} = \left(\mathbf{R}^{h}_{a,t}\right)^{\top} \left(\mathbf{t}^{o}_{t}-\mathbf{p}^{h}_{a,t}\right).$ For anchors belonging to the body joints, we perform forward kinematic to obtain $\mathbf{p}^{h}_{a,t}, \mathbf{R}^{h}_{a,t}$ and for the free anchor we have $\mathbf{p}^{h}_{\text{free},t}=0, \mathbf{R}^{h}_{\text{free},t}=\mathbf{I}$. Hence $\mathbf{t}^{o}_{\text{free},t}=\mathbf{t}^{o}_{t}$. The object motion representation at each frame can hence be written as: $\{\mathbf{R}_t^o, \boldsymbol{\Delta}^{o}_{a,t}, a\in \mathcal{A}\}$.

Our object VAE decoder reconstructs the object rotation $\mathbf{R}_t^o$, the relative translations
$\widehat{\boldsymbol{\Delta}}^{o}_{a,t}$, and routing logits
$\boldsymbol{\alpha}_{t}\in \mathbb{R}^K$ for the $K$ anchors. 
Each body-part-anchored trajectory is then $\widehat{\mathbf{t}}^{o}_{a,t}= \widehat{\mathbf{p}}^{h}_{a,t} +\widehat{\mathbf{R}}^{h}_{a,t}\widehat{\boldsymbol{\Delta}}^{o}_{a,t}.$
The final object translation is obtained through learned soft voting:
\begin{equation}
    \widehat{\mathbf{t}}^{o}_{t}
    =
    \sum_{a\in\mathcal{A}}
    w_{a,t}\widehat{\mathbf{t}}^{o}_{a,t},
    \qquad
    \mathbf{w}_{t}=\operatorname{softmax}(\boldsymbol{\alpha}_{t}).
    \label{eq:anchor-combine}
\end{equation}
At decoding time, the anchor coordinate frames are computed from the decoded human motion, coupling
each trajectory dynamically to the anchors via learned voting. 
We show in \cref{tab:ablation_anchors} that this representation produces better coordinated object motion relative to human.

\textbf{Part-factored human--object interaction encoding.}
We encode human and object motion separately. For human, we represent the motion following \cite{Xiao_2025_ICCV} except we use global rotation $\mathbf{R}^h$ and translation $\mathbf{t}^h$ instead of relative ones, 
and adopt the semantic
body-part partition used in FrankenMotion~\citep{li2026frankenmotion}. Let $\mathcal{B}$ denote the
set of human body parts and $\mathbf{x}^{h,b}_{1:T}$ the motion features of part
$b\in\mathcal{B}$. We introduce a temporal variational encoder $E^{h}_{b}$ for each part, producing $\mathbf{z}^{h,b}=E^{h}_{b}\!\left(\mathbf{x}^{h,b}_{1:T}\right)\in\mathbb{R}^{T'\times d_z}, b\in\mathcal{B}.$
For object motion, a temporal variational encoder $E^{o}$ takes the rigid-object rotation $\mathbf{R}_{1:T}^o$ and anchor relative translations $\boldsymbol{\Delta}^{o}_{a,1:T}$ defined above as input:  $\mathbf{z}^{o}=E^o\left(\mathbf{R}_{1:T}^o, \{\boldsymbol{\Delta}^{o}_{a,1:T}\}_{a\in\mathcal{A}}\right)$. 

All encoders downsample the input by a factor of four, such that each latent token represents four
motion frames and $T'=T/4$. These encoders extract human and object motion features separately, we then apply several layers of spatiotemporal attention to allow them to exchange information and learn the coupling of interactions. Each layer applies self attention with all human-part and object tokens stacked together, followed by causal temporal attention over the sequence. 
Denoting $F_\text{ST}$ as the spatiotemporal layers, the resulting VAE latent is:
\begin{equation}
    \mathbf{Z}
    =
    F_{\mathrm{ST}}\!\left(
    [\mathbf{z}^{h,1},\ldots,\mathbf{z}^{h,|\mathcal{B}|},\mathbf{z}^{o}]
    \right)
    \in
    \mathbb{R}^{T'\times (|\mathcal{B}|+1)\times d_z},
    \label{eq:latent-grid}
\end{equation}
Since the attention layers keep the original token positions, we simply split $\mathbf{Z}$ along the token dimension into
$\mathbf{Z}^{h}\in\mathbb{R}^{T'\times|\mathcal{B}|\times d_z}$, which contains the latent
sequence of every human body part, and $\mathbf{Z}^{o}\in\mathbb{R}^{T'\times d_z}$, which contains
the object latent sequence.

\textbf{Interaction decoding.} To predict interaction motions, we upsample $\mathbf{Z}^{h}, \mathbf{Z}^{o}$ and use separate modules $D_h, D_o$ to decode human and object motions respectively: $\widehat{\mathbf{x}}^{h}_{1:T}=D_h(\mathbf{Z}^{h}),\widehat{\mathbf{x}}^{o}_{1:T}=D_o(\mathbf{Z}^{o};\widehat{\mathbf{x}}^{h}_{1:T})$.
Here, $D_o$ denotes the object decoder prediction followed by the part-anchored composition in
Eq.~\ref{eq:anchor-combine}. 
Architectural details are provided in
Appendix~\ref{app:impl}.

\textbf{Training objective.}
We train the VAE with standard L2 distance between predicted and ground truth items, KL divergence loss $\mathcal{L}_{\mathrm{KL}}$ and auxiliary losses $\mathcal{L}_\text{comp}, \mathcal{L}_w$ to supervise anchor weights prediction:
\begin{equation}
    \mathcal{L}_{\mathrm{VAE}}
    =
    \lambda_h\mathcal{L}^{h}_{\mathrm{rec}}
    +
    \lambda_o\mathcal{L}^{o}_{\mathrm{rec}}
    +
    \lambda_\text{KL}\mathcal{L}_{\mathrm{KL}}
    +
    \lambda_{\mathrm{comp}}\mathcal{L}_{\mathrm{comp}}
    +
    \lambda_{\mathrm{w}}\mathcal{L}_{\mathrm{w}}
    .
    \label{eq:vae-objective}
\end{equation}
where $\mathcal{L}^h_\text{rec}, \mathcal{L}^o_\text{rec}$ are the L2 loss on the reconstructed human and object motions respectively. 
Because these terms do not directly supervise the result after anchor composition, we additionally compare the composed translation from Eq.~\ref{eq:anchor-combine} with the observed object
translation: $\mathcal{L}_{\mathrm{comp}}=\frac{1}{3T}\sum_{t=1}^{T}\left\|\widehat{\mathbf{t}}^{o}_{t}-\mathbf{t}^{o}_{t}\right\|_2^2$.
To warm start and regularize the weight prediction, we also introduce L2 loss between the predicted weights $\mathbf{w}^a_t$ and pseudo ground truth weights $\Tilde{\mathbf{w}}^a_t$. We assign $\Tilde{\mathbf{w}}^a_t$ to one if any vertex of this body part has contact with the object and zero otherwise. For frames with no contacts to any parts, we assign the free anchor weight $\Tilde{\mathbf{w}}^\text{free}_t$ as one and keep others as zero. The weights are then normalized via softmax and contact is defined as distance smaller than 1cm. Hence $\mathcal{L}_w = \frac{1}{T|\mathcal{A}|} \sum_{t=1}^{T} \sum_{a\in \mathcal{A}}\left\|\mathbf{w}_{t}^a-\widetilde{\mathbf{w}}_{t}^a\right\|_2^2.$ We provide details of the loss weights in Appendix.

\subsection{Text-conditioned latent interaction generation}\label{sec:latent-generation}

\vaeName provides a compact latent space for coupled human-object motion. We cast interaction generation as a coarse-to-fine process, first generating coarse interactions via a flow matching model \generatorName to generate latents $\mathbf{Z}$ conditioned on text prompt $c$ and canonical object mesh $\mathcal{M}$. 

\textbf{Rectified-flow training.}
For a clean interaction latent $\mathbf{Z}_0$, we draw
$\tau\sim\mathcal{U}[0,1]$ and
$\boldsymbol{\epsilon}\sim\mathcal{N}(\mathbf{0},\mathbf{I})$, and construct noisy latent via: $\mathbf{Z}_{\tau}=(1-\tau)\mathbf{Z}_0+\tau\boldsymbol{\epsilon},\mathbf{v}^{\star}=\boldsymbol{\epsilon}-\mathbf{Z}_0$. 
The transformer predicts
$\mathbf{v}_{\theta}(\mathbf{Z}_{\tau},\tau,c,\mathcal{M})$ and we train it with standard rectified flow loss $\mathcal{L}_{\mathrm{RF}}$~\cite{liu2023flow}.

\textbf{Explicit Euclidean space motion supervision.}
The rectified-flow objective supervises latent velocity but does not directly constrain the object
trajectories recovered in physical space. We therefore estimate the clean latent and decode it to compute losses defined in explicit Euclidean space. 

The clean latent is computed as: $\widehat{\mathbf{Z}}_0=\mathbf{Z}_{\tau}-\tau\mathbf{v}_{\theta}(\mathbf{Z}_{\tau},\tau,c,\mathcal{M})$.
Decoding this latent yields predictions of human and object motions which we denote as $\hat{\cdot}$. We introduce self-consistency loss $\mathcal{L}_\text{align}$ between object motion $\hat{\mathbf{t}}_{\text{free},t}^\text{o}$ anchored to free trajectory and motions anchored to body parts and Huber distance loss $\mathcal{L}_\text{traj}$ on the predicted object trajectories. Hence the total training loss is: $\mathcal{L}=\lambda_\text{RF}\mathcal{L}_\text{RF} + \lambda_\text{align}\mathcal{L}_\text{align} + \lambda_\text{traj}\mathcal{L}_\text{traj}$,  
where $\mathcal{L}_\text{align}$ forces the predicted free trajectory $\hat{\mathbf{t}}_{\text{free},t}^\text{o}$ to be close to the body part anchored trajectories $\hat{\mathbf{t}}_{a,t}^{o}$: $\mathcal{L}_{\mathrm{align}}=\sum_{t\in\mathcal{C}}\left[\sum_{a\in\mathcal{A}}\overline{w}_{a,t}H_{\delta}(\widehat{\mathbf{t}}^{o}_{\text{free},t},\widehat{\mathbf{t}}^{o}_{a,t})\right]$, where $t\in\mathcal{C}$ denotes the loss is applied only to frames where object is in contact with human and $H_\delta$ denotes Huber distance. We adopt Huber distance to allow stochastic generation of motions that satisfy interaction constrain but deviate from ground truth trajectory. Exact masks, weights, and the noise interval are provided in Appendix~\ref{app:impl}.

\textbf{Sampling.}
At inference, we start from Gaussian noise at $\tau=1$, integrate backward to $\tau=0$ with Classifier-free guidance, and decode the resulting interaction latent. 

\subsection{Recursive geometric refinement}
\label{sec:refinement}
The second stage of our hierarchical generation is fine-grained generation by training \refineName using a hybrid local surface sensing representation to iteratively update the fine interaction details. 

\textbf{Hybrid surface sensors.} Our sensors includes surface geometric probes to reason both long range and short range influence of body to the object for more geometrically correct interaction. We denote $F_{\text{long}, t}, F_{\text{short}, t}$ as the features extracted from long and short probes at frame $t$ respectively and $p^h_{j, t}$ as the SMPL-H skeleton joint at index $j$. Similar to GRAFT~\citep{ym2026graft}, the long range probe features $F_\text{long}$ include the direction and distance from $J_b=22$ body joints to its closest object surface point $p^o_{j,t}$, together corresponding surface normal at $p^o_{j,t}$: $F_{\text{long}, t}=\{(\mathbf{v}_{j,t}, ||\mathbf{v}_{j,t}||_2, \mathbf{n}(p^o_{j,t})), \mathbf{v}_{j,t}=p^o_{j,t}-p^h_{j,t}, \forall j \in \{1,...,J_b\}\}$. These features provide information of how overall body pose affects the object movement but it cannot reason the fine-grained grasp pose which requires more detailed information about the object surface geometry near the fingers. To this end, we introduce short range probes on the hand joints and fingertips that includes not only the direction, distance and object surface normal, but also the object surface points $\mathcal{P}^o_{j, t}$ that are inside a given radius $r$ around these hand points, see visualization in \cref{fig:method} stage 3 topleft. Denoting $J_h$ as the total number of hand joints and fingertips, we have $F_{\text{short}, t}=\{(\mathbf{v}_{j,t}, ||\mathbf{v}_{j,t}||_2, \mathbf{n}(p^o_{j,t}), \mathcal{P}^o_{j, t}), \forall j \in \{1,...,J_h\}\}$ where $\mathcal{P}^o_{j, t}=\{p^o_{t} |\quad \forall p^o_{t} \in \mathcal{M}_t, ||p^o_{t}-p^h_{j,t}||_2<r\}$. Following GEARS~\cite{zhou2024gears}, we transform object points $\mathcal{P}^o_{j, t}$ to the local coordinate of its query point and compute geometric features like distance and surface normals as input to our \refineName. We uniformly sample points on the object surface to obtain $\mathcal{M}_t$ and use PointNet~\cite{qi2017pointnet} like encoder to map them into a fixed dimensional patch feature. Empty spheres are zero-padded. More details can be found in appendix.

\textbf{Refinement network \refineName.} The input to our network includes the hybrid probe features $F_{\text{long}, t}, F_{\text{short}, t}$, human shape parameter, global 6D human pose and object 6D pose. We use an MLP to extract features and then use transformer to compute attention across all tokens and through time. The prediction heads then output delta updates to the coarse human-object interaction motion decoded from generated latents: $\delta_{\theta^h}, \delta_{\mathbf{R}^h}, \delta_{\mathbf{t}^h}, \delta_{\mathbf{R}^o}, \delta_{\mathbf{t}^o}$.

\textbf{Training with mixed data.} We train our network to refine both corrupted GT and generated motions, denoted as \textit{recovery stream} and \textit{generation stream} respectively. This ensures it learns the error pattern of our latent generator while also maintaining generalization. To obtain recovery stream data, we use our trained VAE to encode ground truth interaction motions, and add temporally smooth noise to the decoded motions. 
For these kinds of input, the GT motions are available hence we train directly using L2 distance between refined output and GT motions. More details in \cref{app:refiner_train}.

The generation stream refines samples from the latent generator. Since a text prompt can admit multiple valid motions, direct supervision against a single ground-truth motion is inappropriate. Instead, we optimize geometric objectives targeting motion and geometric quality:
\begin{equation}
    \mathcal{L}_{\mathrm{gen}}
    =\lambda_c\mathcal{L}_{\mathrm{contact}}
    +\lambda_p\mathcal{L}_{\mathrm{penetration}}
    +\lambda_s\mathcal{L}_{\mathrm{support}}
    +\lambda_m\mathcal{L}_{\mathrm{smooth}}
    +\lambda_n\mathcal{L}_{\mathrm{preserve}}.
    \label{eq:refine-real}
\end{equation}
Here, $\mathcal{L}_{\mathrm{contact}}$ pulls relevant body parts toward the object and reduces slip during contact, while $\mathcal{L}_{\mathrm{penetration}}$ penalizes joints inside the object. $\mathcal{L}_{\mathrm{support}}$ prevents planted feet from sliding, drifting, or penetrating the ground; $\mathcal{L}_{\mathrm{smooth}}$ regularizes abrupt human and object motion; and $\mathcal{L}_{\mathrm{preserve}}$ discourages unnecessary changes to non-interacting body regions, the root, and object trajectory. All losses are evaluated at each refinement step and averaged over three iterations.

At inference, the same feed-forward \refineName is applied recursively for a small fixed number of steps. We show in \cref{tab:ablation_refiner} that our mixed data training is important for more faithful generation. More implementation details are provided in
Appendix~\ref{app:impl}.

%% file: sections/experiments.tex
\section{Experiments}
We compare against existing baselines and ablate key components: our method outperforms prior approaches on most metrics, and the ablations confirm the value of each design. \refineName also generalizes zero-shot to interactions produced by other methods, improving their geometry.

\textbf{Experimental and evaluation setup.} We conduct experiments on the standard InterAct~\citep{xu2025interact} benchmark and follow LIGHT~\citep{wang2026unleashing} to report metrics along three dimensions: \textbf{semantic alignment} using R-precision and multimodal distance (MM Dist); \textbf{motion quality} using FID, diversity, and foot-skating ratio (FSR); and \textbf{interaction geometry} using penetration depth (Pene), contact ratio (Contact), and interaction precision, recall, and F1. While LIGHT evaluates the latter three metrics only on hand joints, we report them for both hand and all body joints. Additional metrics are reported in the appendix.

\subsection{Baseline Comparison}
\input{tables/main}

\begin{figure*}[t]
    \centering
    \includegraphics[width=0.99\textwidth]{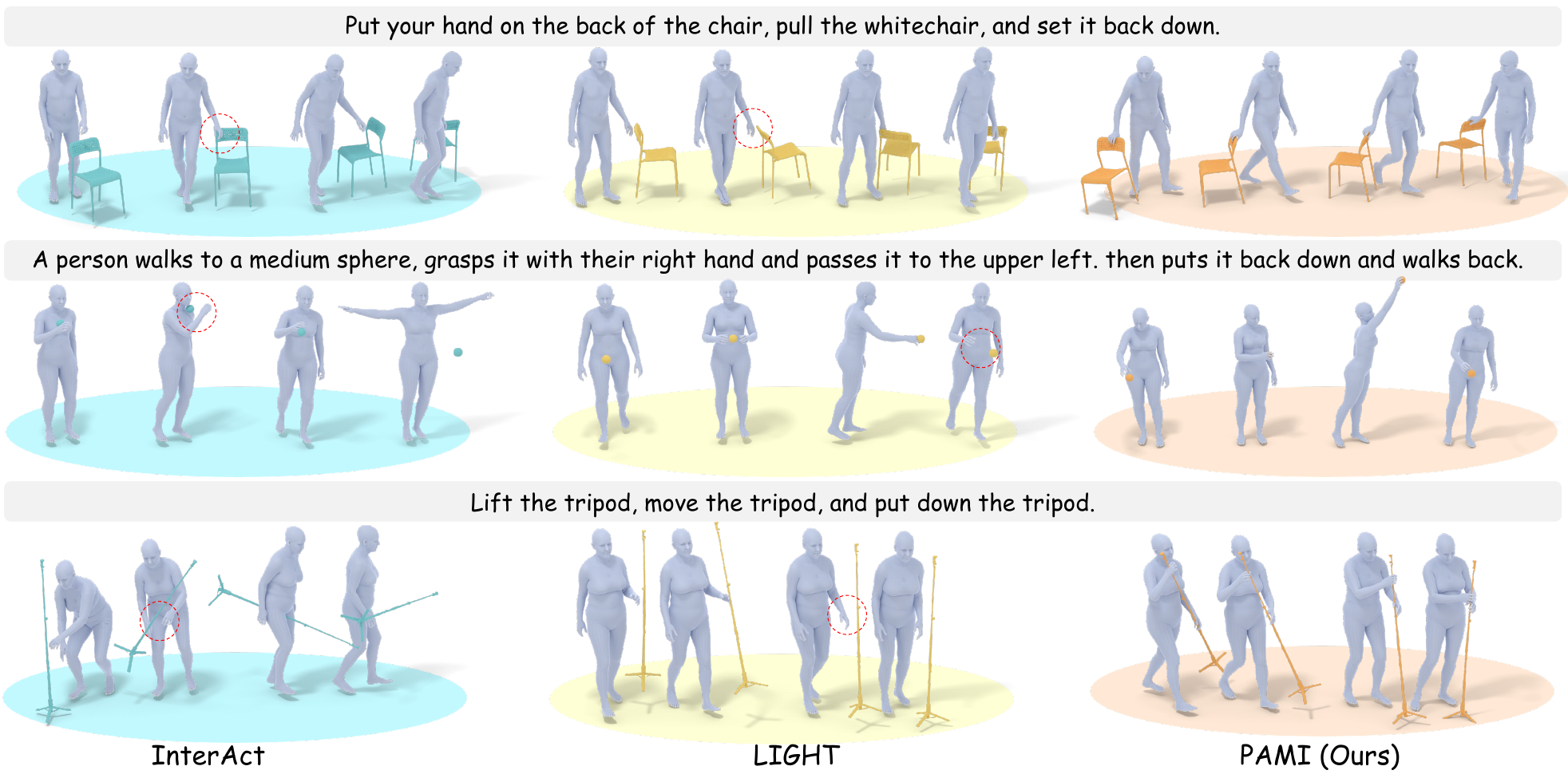}
    \vspace{-10pt}
  \caption{\textbf{Qualitative comparison.} Each row shows four frames from InterAct, LIGHT, and \methodname. InterAct and LIGHT often produce incorrect object orientation or interaction with missing contacts. \methodname better follows the described motion while maintaining coherent contact.}
    \label{fig:qualitative}
    \vspace{-18pt}
\end{figure*}

We compare our method against existing baselines quantitatively in Table~\ref{tab:main} and qualitatively in Figure~\ref{fig:qualitative}. Table~\ref{tab:main} shows that our method achieves large improvements in both semantic alignment (R-1) and interaction geometry, particularly in penetration and
contact.
As shown in Figure~\ref{fig:qualitative}, LIGHT~\citep{wang2026unleashing} models human and object motion as separate trajectories, so the object often fails to follow the human (chair, tripod).
InterAct~\citep{xu2025interact} relies on body-surface markers and their object distances, spreading attention across markers that may be far from the object hence it often misses contact, leaving gaps or unrealistic relative motion.
Our anchor representation directly captures human--object coupling in a compact latent space for coherent generation, while local surface sensing refines interaction geometry.
Even on the slender tripod (third row), our model keeps a secure, natural grasp without penetration.

\subsection{Ablation Studies}
\input{tables/ablation_refined_v2}
We ablate the main components of our method along two dimensions: the part-anchored representation and latent structure in Table~\ref{tab:ablation_anchors}, and the design of \refineName in Table~\ref{tab:ablation_refiner}.

\subsubsection{Part anchored voting representation}\label{subsec:ablate-anchors}
Table~\ref{tab:ablation_anchors}
evaluates the representation designs after refinement.
For a fair comparison, each generator variant is paired with a \refineName trained on its own outputs. The corresponding coarse generations before refinement are reported in Appendix~\ref{tab:ablation}. Together, these evaluations
examine whether the advantages of the proposed representation persist after geometric refinement.

\textbf{Part-factored human encoder.}
We use a continuous part-based VAE that encodes each body part into a separate latent stream and decodes all jointly. Replacing the part encoders with a single full-body encoder, while keeping all else fixed, degrades all metrics (\cref{tab:ablation_anchors}a). The benefit appears already in autoencoder where part factorization lowers reconstruction error by 60.8\% (\cref{tab:vae-recon}).

\textbf{Part-anchored object motion representation.}
Our object representation encodes part-anchored trajectories in the object latent and combines them through learned voting.
We ablate the design choices to understand the importance of anchors, learned weights, and the coordinate representation of global human and object root trajectories. 
The performance degrades gradually after removing non-wrist anchors (\cref{tab:ablation_anchors}b) or all anchors (\cref{tab:ablation_anchors}c), indicating the importance of our anchor representation and reasoning anchors beyond hands. 
Our model predicts dynamic per-frame voting weights to combine anchored predictions, rather than averaging them uniformly (\cref{tab:ablation_anchors}d).
This better models the local influence of each anchor on the object motion, leading to better hand and body contacts.
We also include the absolute poses of the human and object roots in the representation, placing them in the same coordinate system instead of using the commonly adopted frame-relative root representation~\citep{Xiao_2025_ICCV}.
As shown in \cref{tab:ablation_anchors}e, all metrics substantially degrade without absolute pose, as errors accumulate independently for the human and object, leading to divergent trajectories.
See the supplementary website for qualitative examples.

\subsubsection{Fine-grained refinement}\label{subsec:refinement}
\input{tables/refiner} 
\textbf{Refinement quality.}
As shown in Rows (a) and (f) of Table~\ref{tab:ablation_refiner}, applying four \refineName steps substantially improves the interaction geometry of the same coarse generations. It nearly halves penetration, brings Contact closer to the ground-truth distribution, and considerably improves both full-body and hand interaction F1. Meanwhile, R-1 remains stable and FID improves, supporting the role of \refineName as a fine geometric generator that preserves the semantics of the generated interaction. See qualitative examples on the supplementary website.

\textbf{Training with mixed data.}
We train \refineName using controlled corruptions to learn ground-truth contact geometry and generation stream to improve geometry errors from the latent generator. Without either one, the performance degrades (\cref{tab:ablation_refiner}b, c) hence the two streams are complementary.

\textbf{Hybrid surface sensors.} We combine long-range probes for overall body--object relations with short-range probes for precise hand grasping. \cref{tab:ablation_refiner}d, e shows that removing long-range probes hurts overall interaction quality, while removing short-range probes mainly degrades hand contacts.

\textbf{Generalization across HOI generators.} Our \refineName can be used as a general fine interaction reasoner to improve interaction motions from other generators. We convert the outputs of InterAct and LIGHT into compatible input to our \refineName via inverse kinematics and apply the same frozen \refineName. As can be seen in \cref{tab:plugin}, our model improves all baseline generations, showing its transferable capability to reason fine-grained interaction geometries.

\input{tables/plugin}

%% file: tables/main.tex
\begin{table}[!h]
\centering
\caption{\textbf{Quantitative evaluation on InterAct}~\citep{xu2025interact}. R-precision is evaluated with a batch size of 64. $\rightarrow$ indicates that closer to the ground truth is better. $\pm$ indicates a 95\% confidence interval. Our method achieves best performance on almost all metrics.}
\label{tab:main}
\resizebox{\textwidth}{!}{%
\begin{tabular}{@{}lccccccccc@{}}
\toprule
\multirow{2}{*}{Method} &
\multirow{2}{*}{\shortstack{R-Precision\\(Top 1)$^\uparrow$}} &
\multirow{2}{*}{FID$^\downarrow$} &
\multirow{2}{*}{\shortstack{Multimodal\\Dist.$^\downarrow$}} &
\multirow{2}{*}{FSR$^\downarrow$} &
\multirow{2}{*}{Pene$^\downarrow$} &
\multirow{2}{*}{Contact$^\rightarrow$} &
\multicolumn{3}{c}{Interaction (hands)$^\uparrow$} \\
\cmidrule(lr){8-10}
& & & & & & & $C_{prec}$ & $C_{rec}$ & $C_{F1}$ \\
\midrule
Ground Truth &
0.859$^{\pm0.000}$ & 0.000$^{\pm0.000}$ & 1.475$^{\pm0.003}$ & 0.051$^{\pm0.000}$ & 0.060$^{\pm0.000}$ & 0.219$^{\pm0.000}$ & 1.000$^{\pm0.000}$ & 1.000$^{\pm0.000}$ & 1.000$^{\pm0.000}$ \\
\midrule
HOI-Diff &
0.703$^{\pm0.015}$ & 0.698$^{\pm0.017}$ & 2.906$^{\pm0.002}$ & 0.042$^{\pm0.000}$ & 0.092$^{\pm0.004}$ & 0.090$^{\pm0.001}$ & 0.754$^{\pm0.009}$ & 0.531$^{\pm0.007}$ & 0.574$^{\pm0.009}$ \\
CHOIS &
0.717$^{\pm0.005}$ & 0.544$^{\pm0.003}$ & 2.659$^{\pm0.006}$ & 0.072$^{\pm0.001}$ & 0.122$^{\pm0.005}$ & 0.141$^{\pm0.002}$ & 0.823$^{\pm0.003}$ & 0.701$^{\pm0.007}$ & 0.723$^{\pm0.004}$ \\
InterDiff &
0.779$^{\pm0.000}$ & 0.302$^{\pm0.018}$ & 2.364$^{\pm0.004}$ & 0.069$^{\pm0.001}$ & 0.110$^{\pm0.006}$ & 0.150$^{\pm0.001}$ & 0.834$^{\pm0.008}$ & 0.718$^{\pm0.003}$ & 0.741$^{\pm0.004}$ \\
Text2HOI &
0.710$^{\pm0.007}$ & 0.284$^{\pm0.012}$ & 2.551$^{\pm0.011}$ & \textbf{0.035$^{\pm0.001}$} & 0.104$^{\pm0.002}$ & 0.108$^{\pm0.000}$ & 0.840$^{\pm0.010}$ & 0.617$^{\pm0.007}$ & 0.670$^{\pm0.006}$ \\
InterAct &
0.805$^{\pm0.008}$ & 0.280$^{\pm0.016}$ & 2.652$^{\pm0.029}$ & 0.065$^{\pm0.009}$ & 0.119$^{\pm0.028}$ & 0.121$^{\pm0.005}$ & 0.668$^{\pm0.004}$ & 0.552$^{\pm0.011}$ & 0.555$^{\pm0.010}$ \\
LIGHT &
0.719$^{\pm0.003}$ & 0.153$^{\pm0.019}$ & 2.627$^{\pm0.023}$ & 0.050$^{\pm0.001}$ & 0.121$^{\pm0.001}$ & 0.152$^{\pm0.002}$ & 0.832$^{\pm0.003}$ & 0.730$^{\pm0.003}$ & 0.751$^{\pm0.003}$ \\
\midrule
\methodname{} &
\textbf{0.840$^{\pm0.003}$} & \textbf{0.092$^{\pm0.003}$} & \textbf{2.195$^{\pm0.016}$} & 0.078$^{\pm0.001}$ & \textbf{0.062$^{\pm0.001}$} & \textbf{0.193$^{\pm0.001}$} & \textbf{0.843$^{\pm0.002}$} & \textbf{0.836$^{\pm0.006}$} & \textbf{0.821$^{\pm0.004}$} \\
\bottomrule
\end{tabular}}%
\end{table}

%% file: tables/ablation_refined_v2.tex
\begin{table}[t]
 \caption{\textbf{Ablation for part anchored voting representation}. Our part factorization, anchor design with learned routing and absolute root representation all contribute to better generation. 
 }
\label{tab:ablation_anchors}
\begin{center}
\resizebox{\textwidth}{!}{%
\begin{tabular}{@{}lcccccccccc@{}}
\toprule
\multirow{2}{*}{} & \multirow{2}{*}{\shortstack{R-Precision\\(Top 1)$^\uparrow$}} & \multirow{2}{*}{FID$^\downarrow$} & \multirow{2}{*}{Pene$^\downarrow$} & \multirow{2}{*}{Contact$^\rightarrow$} & \multicolumn{3}{c}{Interaction (full body)$^\uparrow$} & \multicolumn{3}{c}{Interaction (hands)$^\uparrow$} \\
\cmidrule(lr){6-8}\cmidrule(lr){9-11}
 &  &  &  &  & $C_{prec}$ & $C_{rec}$ & $C_{F1}$ & $C_{prec}$ & $C_{rec}$ & $C_{F1}$ \\
\midrule
Ground Truth & 0.859 & 0.000 & 0.060 & 0.219 & 1.000 & 1.000 & 1.000 & 1.000 & 1.000 & 1.000 \\
\midrule
Our full \methodname model & \textbf{0.840$^{\pm0.003}$} & \textbf{0.092$^{\pm0.003}$} & \textbf{0.062$^{\pm0.001}$} & \textbf{0.193$^{\pm0.001}$} & \textbf{0.502$^{\pm0.003}$} & \textbf{0.477$^{\pm0.003}$} & \textbf{0.457$^{\pm0.004}$} & 0.843$^{\pm0.002}$ & \textbf{0.836$^{\pm0.006}$} & \textbf{0.821$^{\pm0.004}$} \\
\midrule
(a) w/o part factorization & 0.822$^{\pm0.002}$ & 0.171$^{\pm0.006}$ & 0.086$^{\pm0.001}$ & 0.129$^{\pm0.002}$ & 0.466$^{\pm0.002}$ & 0.294$^{\pm0.005}$ & 0.327$^{\pm0.004}$ & 0.823$^{\pm0.007}$ & 0.657$^{\pm0.006}$ & 0.697$^{\pm0.006}$ \\
(b) w/o leg and root anchors & \textbf{0.840$^{\pm0.002}$} & 0.097$^{\pm0.006}$ & 0.087$^{\pm0.002}$ & 0.174$^{\pm0.001}$ & 0.499$^{\pm0.005}$ & 0.425$^{\pm0.004}$ & 0.425$^{\pm0.004}$ & \textbf{0.849$^{\pm0.002}$} & 0.806$^{\pm0.004}$ & 0.804$^{\pm0.003}$ \\
(c) w/o anchors & 0.834$^{\pm0.003}$ & 0.102$^{\pm0.004}$ & 0.076$^{\pm0.003}$ & 0.116$^{\pm0.001}$ & 0.478$^{\pm0.015}$ & 0.277$^{\pm0.007}$ & 0.313$^{\pm0.009}$ & 0.830$^{\pm0.006}$ & 0.647$^{\pm0.003}$ & 0.686$^{\pm0.001}$ \\
(d) w/o learned voting weights & 0.831$^{\pm0.001}$ & 0.099$^{\pm0.005}$ & 0.089$^{\pm0.003}$ & 0.175$^{\pm0.002}$ & 0.495$^{\pm0.005}$ & 0.423$^{\pm0.002}$ & 0.423$^{\pm0.002}$ & 0.844$^{\pm0.003}$ & 0.807$^{\pm0.009}$ & 0.802$^{\pm0.007}$ \\
(e) w/o absolute root & 0.780$^{\pm0.005}$ & 0.401$^{\pm0.011}$ & 0.063$^{\pm0.002}$ & 0.126$^{\pm0.002}$ & 0.478$^{\pm0.006}$ & 0.312$^{\pm0.005}$ & 0.342$^{\pm0.003}$ & 0.829$^{\pm0.006}$ & 0.698$^{\pm0.002}$ & 0.720$^{\pm0.002}$ \\
\bottomrule
\end{tabular}}
\end{center}
\end{table}

%% file: tables/refiner.tex
\begin{table}[t]
\caption{\textbf{\refineName ablations}. All variants refine the same coarse generations. Mixed data training and hybrid sensing are important to obtain high-quality fine interaction details.}
\label{tab:ablation_refiner}
\begin{center}
\resizebox{\textwidth}{!}{%
\begin{tabular}{@{}lcccccccccc@{}}
\toprule
\multirow{2}{*}{} & \multirow{2}{*}{\shortstack{R-Precision\\(Top 1)$^\uparrow$}} & \multirow{2}{*}{FID$^\downarrow$} & \multirow{2}{*}{Pene$^\downarrow$} & \multirow{2}{*}{Contact$^\rightarrow$} & \multicolumn{3}{c}{Interaction (full body)$^\uparrow$} & \multicolumn{3}{c}{Interaction (hands)$^\uparrow$} \\
\cmidrule(lr){6-8}\cmidrule(lr){9-11}
 &  &  &  &  & $C_{prec}$ & $C_{rec}$ & $C_{F1}$ & $C_{prec}$ & $C_{rec}$ & $C_{F1}$ \\
\midrule
Ground Truth & 0.859 & 0.000 & 0.060 & 0.219 & 1.000 & 1.000 & 1.000 & 1.000 & 1.000 & 1.000 \\
\midrule
(a) no refinement & 0.839$^{\pm0.004}$ & 0.103$^{\pm0.004}$ & 0.115$^{\pm0.003}$ & 0.142$^{\pm0.001}$ & 0.500$^{\pm0.006}$ & 0.364$^{\pm0.005}$ & 0.389$^{\pm0.004}$ & \textbf{0.849$^{\pm0.001}$} & 0.758$^{\pm0.003}$ & 0.775$^{\pm0.002}$ \\
\midrule
(b) w/o generation stream & \textbf{0.841$^{\pm0.003}$} & 0.101$^{\pm0.005}$ & 0.091$^{\pm0.001}$ & 0.155$^{\pm0.000}$ & 0.496$^{\pm0.003}$ & 0.388$^{\pm0.002}$ & 0.404$^{\pm0.001}$ & 0.847$^{\pm0.002}$ & 0.788$^{\pm0.003}$ & 0.794$^{\pm0.002}$ \\
(c) w/o corruption stream & \textbf{0.841$^{\pm0.003}$} & 0.104$^{\pm0.003}$ & 0.085$^{\pm0.003}$ & 0.166$^{\pm0.001}$ & 0.496$^{\pm0.005}$ & 0.414$^{\pm0.004}$ & 0.419$^{\pm0.002}$ & 0.847$^{\pm0.005}$ & 0.786$^{\pm0.004}$ & 0.793$^{\pm0.004}$ \\
(d) w/o short-range probes & 0.840$^{\pm0.003}$ & 0.100$^{\pm0.004}$ & 0.083$^{\pm0.002}$ & 0.186$^{\pm0.000}$ & 0.496$^{\pm0.002}$ & 0.462$^{\pm0.001}$ & 0.446$^{\pm0.000}$ & 0.845$^{\pm0.002}$ & 0.828$^{\pm0.003}$ & 0.816$^{\pm0.002}$ \\
(e) w/o long-range probes & 0.840$^{\pm0.003}$ & 0.101$^{\pm0.004}$ & 0.111$^{\pm0.003}$ & 0.137$^{\pm0.000}$ & 0.500$^{\pm0.007}$ & 0.351$^{\pm0.003}$ & 0.379$^{\pm0.004}$ & \textbf{0.849$^{\pm0.007}$} & 0.750$^{\pm0.003}$ & 0.770$^{\pm0.003}$ \\
\midrule
(f) \refineName & 0.840$^{\pm0.003}$ & \textbf{0.092$^{\pm0.003}$} & \textbf{0.062$^{\pm0.001}$} & \textbf{0.193$^{\pm0.001}$} & \textbf{0.502$^{\pm0.003}$} & \textbf{0.477$^{\pm0.003}$} & \textbf{0.457$^{\pm0.004}$} & 0.843$^{\pm0.002}$ & \textbf{0.836$^{\pm0.006}$} & \textbf{0.821$^{\pm0.004}$} \\
\bottomrule
\end{tabular}}
\end{center}
\end{table}

%% file: tables/plugin.tex
\begin{figure}[t]
    \centering
    \begin{minipage}[t]{0.49\columnwidth}
        \centering
        \vspace{0pt}
        \includegraphics[width=\linewidth]{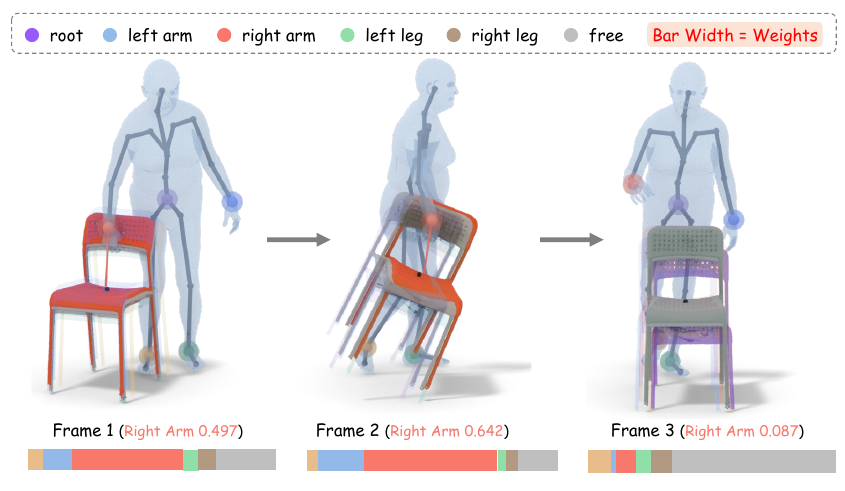}
        \captionof{figure}{\textbf{Body-part anchor weights} strongly correlate with contact body parts.}
        \label{fig:case_study}
    \end{minipage}
    \hfill
    \begin{minipage}[t]{0.49\columnwidth}
        \centering
        \vspace{0pt}
        \captionof{table}{\textbf{Generalization of \refineName} to LIGHT and InterAct. Full metrics are in Table~\ref{tab:plugin_supp}.
        }
        \label{tab:plugin}

        \small
        \setlength{\tabcolsep}{2.5pt}
        \begin{tabular*}{\linewidth}{@{\extracolsep{\fill}}lcccc@{}}
        \toprule
         & Pene$^\downarrow$ & Contact$^\rightarrow$
         & F1$_{\text{body}}^{\uparrow}$ & F1$_{\text{hand}}^{\uparrow}$ \\
        \midrule
        Ground Truth & 0.060 & 0.219 & 1.000 & 1.000 \\
        \midrule
        LIGHT & 0.134 & 0.128 & 0.328 & 0.704 \\
        \quad $+$ w/o corr. & 0.094 & 0.151 & 0.361 & 0.712 \\
        \quad $+$ full & \textbf{0.069} & \textbf{0.181} & \textbf{0.408} & \textbf{0.755} \\
        \midrule
        InterAct & 0.126 & 0.122 & 0.278 & 0.658 \\
        \quad $+$ w/o corr. & 0.098 & 0.150 & 0.315 & 0.687 \\
        \quad $+$ full & \textbf{0.077} & \textbf{0.188} & \textbf{0.374} & \textbf{0.738} \\
        \bottomrule
        \end{tabular*}
    \end{minipage}
    \vspace{-14pt}
\end{figure}

%% file: sections/conclusion.tex
\vspace{-8pt}
\section{Conclusion and limitations}
We presented \methodname for text-conditioned full-body HOI generation. Following the Hough-Transform idea, we represent object motion as votes from body-part anchors and learn per-frame weights that aggregate the relevant votes, and we generate the interaction in a compact part-based latent space. Generation is coarse-to-fine: \generatorName produces a coarse interaction in this latent space, and \refineName then refines local interaction poses from hybrid surface geometry sensing. On InterAct, \methodname achieves best performance in almost all metrics, achieving 14.5\% higher contact recall than the previous art, and our ablations confirm that the part-based latent, anchor voting, learned weights, and \refineName each contribute. \refineName further improves LIGHT and InterAct outputs without retraining. Our code and models will be publicly released.

%% file: sections/appendix.tex
\clearpage

\makeatletter
\renewcommand{\partname}{}
\renewcommand{\thepart}{}
\makeatother

\part{Appendix} 
{
  \hypersetup{linkcolor=black}
  \parttoc
}
\suppressfloats[t]

\section{Additional experiments}
Table~\ref{tab:main-supp} reports the retrieval and diversity metrics omitted from the main table.
\input{tables/main_supp}

\subsection{Number of refinement steps}
We vary the number of recursive updates at inference to select the refinement depth.
\begin{figure}[htbp]
    \centering
    \includegraphics[width=0.95\linewidth]{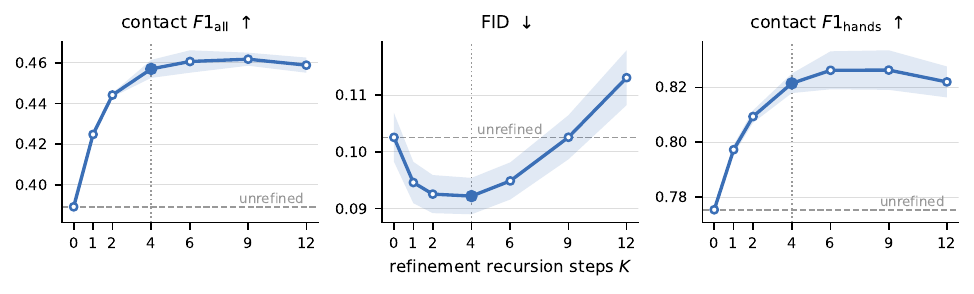}
    \caption{\textbf{Effect of recursion depth $K$.} From left to right: full-body contact F1, FID,
    and hand-contact F1. Curves show the mean over three generation draws; bands denote 95\%
    confidence intervals.}
    \label{fig:kcurve}
\end{figure}

Most contact improvement occurs within four updates. Further refinement gives little contact gain
and eventually worsens FID. We therefore use $K{=}4$ throughout.

\subsection{Runtime and memory}
\vaeName, \generatorName, and \refineName have 755M, 55M, and 4.4M parameters, respectively. On one
H100, \generatorName trains for 240k iterations in about 16 hours and \refineName for 44k iterations
in about 8 hours. For a 231-frame sequence at batch size 1, generation takes 0.13\,s and 5.7\,GB;
four refinement steps add 0.66\,s and use 2.8\,GB.

\section{Implementation details}\label{app:impl}

\subsection{Motion representation and coordinates}
Motion is processed at 30 fps in windows of at most 300 frames. For each sequence, the human root
in the first frame defines the world coordinate frame, and all subsequent global human and object
motion is expressed relative to this frame. We follow MotionStreamer~\citep{Xiao_2025_ICCV} for the
272-dimensional body representation, but replace its frame-relative root variables with global root
orientation and translation in this first-frame coordinate system. The remaining features contain
root-relative joint positions and velocities, and parent-relative 6D rotations. The two hands are
represented separately by wrist-local positions and parent-relative 6D rotations for 15 finger
joints per side.

\vaeName uses eight causal body-part encoders for the root, torso, left/right arms, legs, and hands.
The object encoder receives its global 6D rotation, free translation, and anchor-relative offsets.
All encoders downsample time by four and produce 16-dimensional tokens. Six spatiotemporal-attention
layers exchange information across tokens and time. One joint human decoder fuses all body-part
tokens, and two upsampling stages restore the original frame rate.
Table~\ref{tab:vae-recon} compares VAE reconstruction quality, while
Table~\ref{tab:ablation} reports the corresponding generators before refinement.
\input{tables/vae_recon}
\input{tables/ablation}

\subsection{Anchor voting implementation}
The representation uses five body-part anchors---the root, left/right wrists, and left/right
ankles---and one free anchor. Each body-part vote transforms a decoded relative offset through its
decoded human anchor; the free vote predicts object translation directly.

The voting target $\widetilde{\mathbf{w}}_t$ assigns weight to body parts within 1 cm of the object.
Multiple active votes are normalized; when no body-part contact is detected, the free branch
receives the full vote. We use
$\mathcal{L}_{\mathrm{w}}=\operatorname{MSE}(\mathbf{w}_t,\widetilde{\mathbf{w}}_t)$ with weight
$0.01$. At inference, the decoder predicts the votes and softmax weights without the target. Object
orientation is decoded separately in 6D.
Table~\ref{tab:ablation_refined_supp} reports the additional metrics for the refined representation
ablations.
\input{tables/ablation_refined_supp}

\subsection{\generatorName: Text-Conditioned Latent Interaction Generation}
The latent grid contains eight human-part tokens and one object token per latent
step. The transformer has 10 layers, width 512, eight attention heads, and feed-forward width 2048.
It applies self-attention to the full token grid and cross-attention to frozen caption features.
Rotary embeddings encode time, while token embeddings distinguish body parts and the object. The
object token is additionally conditioned on basis-point-set features of the canonical
mesh~\citep{prokudin2019efficient}. All normalization statistics are computed on the training split;
human statistics are computed separately for each part token. 
To prevent the object loss from being
diluted by the larger number of human-part tokens, we compute the human and object errors
separately:
\begin{equation}
    \mathcal{L}_{\mathrm{RF}}
    =
    \mathbb{E}_{\tau,\boldsymbol{\epsilon}}
    \left[
    \left\|
    \mathbf{v}^{h}_{\theta}-\mathbf{v}^{\star,h}
    \right\|_2^2
    +
    \lambda_o
    \left\|
    \mathbf{v}^{o}_{\theta}-\mathbf{v}^{\star,o}
    \right\|_2^2
    \right].
    \label{eq:rf}
\end{equation}

We use classifier-free guidance with scale $2.5$ and 50 integration steps. Decoded losses use Huber
distance and are applied only for $0.1<\tau<0.5$. The free-position, free-velocity, and composed
translation terms each have weight $5.0$; alignment has weight $2.5$. Alignment is evaluated only
on frames with an active body-part vote, after removing the predicted free weight and renormalizing
the remaining weights. Targets come from decoding the clean latent with the same VAE.

\subsection{\refineName architecture and rollout}
For one object, \refineName uses 54 tokens per frame: 22 body joints, 30 finger joints, one object,
and one global token. Joint tokens contain 6D rotation and long-range surface probes. Body probes
use the root frame, while finger probes use the wrist frame. The object token contains object pose;
the global token contains root pose and a 16D body-shape feature. Long-range probes use 512 surface
points.

Short-range probes use a separate 16,384-point area-weighted surface sample. For each finger joint
and fingertip, we encode at most 64 points within 4\,cm using a shared pointwise network and
masked-mean pooling. Empty neighborhoods produce zero features. Both probe types are recomputed
after every refinement step.

Features are projected to width 256 and processed by four spatial--temporal transformer blocks.
Zero-initialized heads predict bounded updates to joint rotations, root motion, and object
$\mathrm{SE}(3)$. Training backpropagates through a three-step rollout. Object updates are capped at
5 cm and $10^\circ$; root updates at 10 cm and $5^\circ$ in yaw.
Table~\ref{tab:plugin_supp} reports the complete cross-generator transfer metrics.
\input{tables/plugin_supp}

\subsection{\refineName training losses}\label{app:refiner_train}
Recovery and generation batches alternate $1{:}1$ while the VAE and generator remain frozen. A
recovery input is a corrupted VAE round trip, supervised by the original interaction. We sample
clean, root, torso, upper-body, lower-body, object, and mixed groups with probabilities
$0.2/0.1/0.1/0.2/0.1/0.2/0.1$; mixed corruption combines two or three groups.

Temporal noise follows an Ornstein--Uhlenbeck process with $\alpha\approx0.895$. Root noise is
4.7 cm in $xz$, $3^\circ$ in yaw, and a 4 cm height offset. Rotation ranges are $2$--$5^\circ$
for the torso, $3$--$7^\circ$ for hips/knees, $2$--$5^\circ$ for ankles, $3$--$12^\circ$ for the
upper body, and $4$--$15^\circ$ for fingers. Object noise uses $1$--$4$ cm translation and
$3$--$12^\circ$ rotation. Persistent biases are $1$--$6^\circ$ for the upper body,
$2$--$8^\circ$ for fingers, and $1$--$5$ cm/$3$--$15^\circ$ for the object. The corruption
distribution is fixed across the three training steps; generated samples are precomputed.

The implementation terms follow the five groups in Eq.~\ref{eq:refine-real}.
$\mathcal{L}_{\mathrm{contact}}$ contains surface attraction and contact slip;
$\mathcal{L}_{\mathrm{penetration}}$ penalizes human--object intersection;
$\mathcal{L}_{\mathrm{support}}$ constrains planted feet, ground penetration, and lower-body drift;
$\mathcal{L}_{\mathrm{smooth}}$ penalizes abrupt motion and updates; and
$\mathcal{L}_{\mathrm{preserve}}$ protects non-interacting body regions, the root, and the object.
The five group weights $\lambda_c,\lambda_p,\lambda_s,\lambda_m,\lambda_n$ are all 1; their internal
coefficients and the separate recovery-stream weights are reported in Table~\ref{tab:hparams}.
Contact masks and segment anchors are computed once from the coarse generation and kept fixed
during recursion.
Table~\ref{tab:refiner_supp} reports the additional metrics for the RefineNet ablations.
\input{tables/refiner_supp}

Parameter counts, optimizer schedules, and the scalar loss weights are listed in
Table~\ref{tab:hparams}, read directly from the final checkpoints' recorded configurations.
\input{tables/hparams}

\section{Limitations and future work}
Our experiments focus on the single-human, single-rigid-object setting of InterAct, and the method assumes access to a known canonical object mesh for shape conditioning and surface sensing, which may limit its applicability when accurate object geometry is unavailable. Moreover, \refineName is designed as a bounded local corrector: it can improve contact when the coarse interaction is already reasonable, but cannot recover from an incorrect action, severe temporal misalignment, or a large trajectory error. Future work will extend the framework to interactions involving multiple objects and improve its generalization to novel object shapes beyond those observed during training.

%% file: tables/main_supp.tex
\begin{table}[h]
\centering
\caption{\textbf{Additional metrics on InterAct.} Top-2/3 R-precision and Diversity complement
Table~\ref{tab:main}.}
\label{tab:main-supp}
\begin{tabular}{@{}lccc@{}}
\toprule
Method & R-Prec. Top 2$^\uparrow$ & R-Prec. Top 3$^\uparrow$ & Diversity$^\rightarrow$ \\
\midrule
Ground Truth & 0.967$^{\pm0.001}$ & 0.992$^{\pm0.002}$ & 7.764$^{\pm0.020}$ \\
\midrule
HOI-Diff & 0.881$^{\pm0.012}$ & 0.929$^{\pm0.003}$ & 7.577$^{\pm0.054}$ \\
CHOIS & 0.880$^{\pm0.004}$ & 0.934$^{\pm0.002}$ & 7.777$^{\pm0.046}$ \\
InterDiff & 0.930$^{\pm0.002}$ & 0.968$^{\pm0.005}$ & 7.738$^{\pm0.058}$ \\
Text2HOI & 0.869$^{\pm0.005}$ & 0.926$^{\pm0.000}$ & 7.730$^{\pm0.009}$ \\
LIGHT & 0.877$^{\pm0.007}$ & 0.930$^{\pm0.001}$ & 7.737$^{\pm0.062}$ \\
InterAct & 0.935$^{\pm0.007}$ & 0.967$^{\pm0.006}$ & \textbf{7.775$^{\pm0.026}$} \\
\midrule
\methodname & \textbf{0.958$^{\pm0.002}$} & \textbf{0.980$^{\pm0.001}$} & 7.693$^{\pm0.007}$ \\
\bottomrule
\end{tabular}
\end{table}

%% file: tables/vae_recon.tex
\begin{table}[t]
\centering
\caption{\textbf{\vaeName reconstruction.} Validation errors from the final checkpoint of each
variant. Body and hand errors are in mm; object translation and rotation errors are in cm and
degrees.}
\label{tab:vae-recon}
\resizebox{\linewidth}{!}{%
\begin{tabular}{@{}lccccc@{}}
\toprule
Variant & MPJPE (mm)$\downarrow$ & hand (mm)$\downarrow$ & object comp.\ (cm)$\downarrow$ & object pos.\ (cm)$\downarrow$ & object rot.\ ($^\circ$)$\downarrow$ \\
\midrule
part-factored (ours) & 15.2 & 1.3 & 2.54 & 1.66 & 2.2 \\
$+$ contact-tuned decoder (final) & 16.7 & 1.3 & 2.49 & 1.26 & 1.6 \\
\midrule
whole-body, param.-matched & 38.8 & 2.3 & 2.70 & 1.38 & 2.0 \\
wrist-only anchors & 16.5 & 1.3 & 2.59 & 1.02 & 2.2 \\
no anchors & 17.7 & 1.2 & -- & 1.41 & 1.2 \\
w/o spatiotemporal attention & 15.4 & 1.2 & 2.94 & 2.10 & 2.5 \\
\bottomrule
\end{tabular}}
\end{table}

%% file: tables/ablation.tex
\begin{table}[t]
\centering
\caption{\textbf{Generator ablations before refinement} ($K{=}0$). These coarse outputs isolate
the representation choices from geometric refinement. We report the mean and 95\% confidence
interval over three generation draws.}
\label{tab:ablation}
\resizebox{\linewidth}{!}{%
\begin{tabular}{@{}lccccccc@{}}
\toprule
 & Top-1$^\uparrow$ & Top-2$^\uparrow$ & Top-3$^\uparrow$ & FID$^\downarrow$ & MM Dist$^\downarrow$ & Diversity$^\rightarrow$ & FSR$^\downarrow$ \\
\midrule
Ground Truth & 0.859 & 0.967 & 0.992 & 0.000 & 1.475 & 7.764 & 0.051 \\
\midrule
Our full \methodname model & 0.839$^{\pm0.004}$ & 0.957$^{\pm0.002}$ & 0.980$^{\pm0.002}$ & \textbf{0.103$^{\pm0.004}$} & \textbf{2.198$^{\pm0.015}$} & 7.683$^{\pm0.008}$ & 0.137$^{\pm0.002}$ \\
\midrule
(a) w/o part factorization & 0.816$^{\pm0.002}$ & 0.946$^{\pm0.001}$ & 0.972$^{\pm0.002}$ & 0.330$^{\pm0.009}$ & 2.375$^{\pm0.004}$ & 7.535$^{\pm0.016}$ & 0.262$^{\pm0.001}$ \\
(b) w/o leg and root anchors & \textbf{0.840$^{\pm0.001}$} & \textbf{0.960$^{\pm0.003}$} & \textbf{0.984$^{\pm0.001}$} & 0.109$^{\pm0.006}$ & 2.215$^{\pm0.014}$ & 7.706$^{\pm0.006}$ & 0.137$^{\pm0.002}$ \\
(c) w/o anchors & 0.832$^{\pm0.004}$ & 0.953$^{\pm0.003}$ & 0.977$^{\pm0.001}$ & 0.119$^{\pm0.007}$ & 2.230$^{\pm0.004}$ & \textbf{7.723$^{\pm0.013}$} & \textbf{0.128$^{\pm0.002}$} \\
(d) w/o learned voting weights & 0.830$^{\pm0.003}$ & 0.950$^{\pm0.001}$ & 0.974$^{\pm0.001}$ & 0.116$^{\pm0.006}$ & 2.237$^{\pm0.007}$ & 7.689$^{\pm0.017}$ & 0.135$^{\pm0.001}$ \\
(e) w/o absolute root & 0.775$^{\pm0.002}$ & 0.909$^{\pm0.006}$ & 0.948$^{\pm0.005}$ & 0.368$^{\pm0.026}$ & 2.684$^{\pm0.042}$ & 7.418$^{\pm0.015}$ & 0.147$^{\pm0.001}$ \\
\bottomrule
\end{tabular}}\\[6pt]
\resizebox{\linewidth}{!}{%
\begin{tabular}{@{}lcccccccc@{}}
\toprule
\multirow{2}{*}{} & \multirow{2}{*}{Pene$^\downarrow$} & \multirow{2}{*}{Contact$^\rightarrow$} & \multicolumn{3}{c}{Interaction (full body)$^\uparrow$} & \multicolumn{3}{c}{Interaction (hands)$^\uparrow$} \\
\cmidrule(lr){4-6}\cmidrule(lr){7-9}
 &  &  & $C_{prec}$ & $C_{rec}$ & $C_{F1}$ & $C_{prec}$ & $C_{rec}$ & $C_{F1}$ \\
\midrule
Ground Truth & 0.060 & 0.219 & 1.000 & 1.000 & 1.000 & 1.000 & 1.000 & 1.000 \\
\midrule
Our full \methodname model & 0.115$^{\pm0.003}$ & \textbf{0.142$^{\pm0.001}$} & \textbf{0.500$^{\pm0.006}$} & \textbf{0.364$^{\pm0.005}$} & \textbf{0.389$^{\pm0.004}$} & \textbf{0.849$^{\pm0.001}$} & \textbf{0.758$^{\pm0.003}$} & \textbf{0.775$^{\pm0.002}$} \\
\midrule
(a) w/o part factorization & \textbf{0.083$^{\pm0.002}$} & 0.069$^{\pm0.001}$ & 0.434$^{\pm0.005}$ & 0.151$^{\pm0.004}$ & 0.199$^{\pm0.003}$ & 0.813$^{\pm0.002}$ & 0.447$^{\pm0.001}$ & 0.532$^{\pm0.002}$ \\
(b) w/o leg and root anchors & 0.115$^{\pm0.003}$ & 0.140$^{\pm0.002}$ & 0.489$^{\pm0.003}$ & 0.353$^{\pm0.005}$ & 0.375$^{\pm0.004}$ & 0.848$^{\pm0.003}$ & 0.754$^{\pm0.004}$ & 0.773$^{\pm0.003}$ \\
(c) w/o anchors & 0.100$^{\pm0.006}$ & 0.091$^{\pm0.001}$ & 0.451$^{\pm0.012}$ & 0.210$^{\pm0.006}$ & 0.254$^{\pm0.009}$ & 0.828$^{\pm0.005}$ & 0.575$^{\pm0.005}$ & 0.633$^{\pm0.003}$ \\
(d) w/o learned voting weights & 0.115$^{\pm0.004}$ & 0.132$^{\pm0.001}$ & 0.485$^{\pm0.003}$ & 0.316$^{\pm0.003}$ & 0.350$^{\pm0.001}$ & 0.848$^{\pm0.000}$ & 0.725$^{\pm0.008}$ & 0.754$^{\pm0.006}$ \\
(e) w/o absolute root & 0.112$^{\pm0.002}$ & 0.115$^{\pm0.002}$ & 0.469$^{\pm0.003}$ & 0.290$^{\pm0.002}$ & 0.323$^{\pm0.004}$ & 0.826$^{\pm0.007}$ & 0.660$^{\pm0.003}$ & 0.694$^{\pm0.007}$ \\
\bottomrule
\end{tabular}}
\end{table}

%% file: tables/ablation_refined_supp.tex
\begin{table}[t]
\centering
\caption{\textbf{Additional metrics for the refined representation ablations in
Table~\ref{tab:ablation_anchors}} ($K{=}4$). Mean and 95\% confidence interval over three runs.}
\label{tab:ablation_refined_supp}
\resizebox{\linewidth}{!}{%
\begin{tabular}{@{}lccccc@{}}
\toprule
 & Top-2$^\uparrow$ & Top-3$^\uparrow$ & MM Dist$^\downarrow$ & Diversity$^\rightarrow$ & FSR$^\downarrow$ \\
\midrule
Ground Truth & 0.967 & 0.992 & 1.475 & 7.764 & 0.051 \\
\midrule
Our full \methodname model & 0.958$^{\pm0.002}$ & 0.980$^{\pm0.001}$ & \textbf{2.195$^{\pm0.016}$} & 7.693$^{\pm0.007}$ & \textbf{0.078$^{\pm0.001}$} \\
\midrule
(a) w/o part factorization & 0.948$^{\pm0.000}$ & 0.973$^{\pm0.001}$ & 2.308$^{\pm0.003}$ & 7.651$^{\pm0.019}$ & 0.115$^{\pm0.001}$ \\
(b) w/o leg and root anchors & \textbf{0.960$^{\pm0.002}$} & \textbf{0.985$^{\pm0.001}$} & 2.207$^{\pm0.013}$ & 7.718$^{\pm0.004}$ & 0.112$^{\pm0.001}$ \\
(c) w/o anchors & 0.953$^{\pm0.002}$ & 0.976$^{\pm0.001}$ & 2.218$^{\pm0.003}$ & \textbf{7.727$^{\pm0.010}$} & 0.092$^{\pm0.001}$ \\
(d) w/o learned voting weights & 0.950$^{\pm0.002}$ & 0.974$^{\pm0.001}$ & 2.228$^{\pm0.006}$ & 7.695$^{\pm0.014}$ & 0.103$^{\pm0.001}$ \\
(e) w/o absolute root & 0.916$^{\pm0.008}$ & 0.951$^{\pm0.009}$ & 2.705$^{\pm0.038}$ & 7.384$^{\pm0.026}$ & 0.096$^{\pm0.001}$ \\
\bottomrule
\end{tabular}}
\end{table}

%% file: tables/plugin_supp.tex
\begin{table}[t]
\caption{\textbf{Full metrics for Table~\ref{tab:plugin}.} Zero-shot refinement of LIGHT and
InterAct outputs with $K{=}4$; mean and 95\% confidence interval over two sampling campaigns. Bold
marks the best result in each generator block.}
\label{tab:plugin_supp}
\begin{center}
\resizebox{\textwidth}{!}{%
\begin{tabular}{@{}lccccccccccccccc@{}}
\toprule
\multirow{2}{*}{} & \multirow{2}{*}{Top-1$^\uparrow$} & \multirow{2}{*}{Top-2$^\uparrow$} & \multirow{2}{*}{Top-3$^\uparrow$} & \multirow{2}{*}{FID$^\downarrow$} & \multirow{2}{*}{MM Dist$^\downarrow$} & \multirow{2}{*}{Diversity$^\rightarrow$} & \multirow{2}{*}{FSR$^\downarrow$} & \multirow{2}{*}{Pene$^\downarrow$} & \multirow{2}{*}{Contact$^\rightarrow$} & \multicolumn{3}{c}{Interaction (full body)$^\uparrow$} & \multicolumn{3}{c}{Interaction (hands)$^\uparrow$} \\
\cmidrule(lr){11-13}\cmidrule(lr){14-16}
 &  &  &  &  &  &  &  &  &  & $C_{prec}$ & $C_{rec}$ & $C_{F1}$ & $C_{prec}$ & $C_{rec}$ & $C_{F1}$ \\
\midrule
Ground Truth & 0.859 & 0.967 & 0.992 & 0.000 & 1.475 & 7.764 & 0.051 & 0.060 & 0.219 & 1.000 & 1.000 & 1.000 & 1.000 & 1.000 & 1.000 \\
\midrule
LIGHT & 0.763$^{\pm0.013}$ & 0.911$^{\pm0.009}$ & \textbf{0.948$^{\pm0.004}$} & \textbf{0.225$^{\pm0.073}$} & 2.643$^{\pm0.013}$ & 7.593$^{\pm0.012}$ & 0.105$^{\pm0.003}$ & 0.134$^{\pm0.005}$ & 0.128$^{\pm0.000}$ & 0.468$^{\pm0.001}$ & 0.308$^{\pm0.004}$ & 0.328$^{\pm0.000}$ & \textbf{0.808$^{\pm0.003}$} & 0.686$^{\pm0.013}$ & 0.704$^{\pm0.010}$ \\
LIGHT $+$ \refineName{} w/o corruption stream & 0.762$^{\pm0.009}$ & 0.910$^{\pm0.009}$ & \textbf{0.948$^{\pm0.002}$} & 0.241$^{\pm0.068}$ & 2.644$^{\pm0.011}$ & \textbf{7.606$^{\pm0.022}$} & 0.061$^{\pm0.001}$ & 0.094$^{\pm0.003}$ & 0.151$^{\pm0.000}$ & 0.459$^{\pm0.001}$ & 0.360$^{\pm0.006}$ & 0.361$^{\pm0.001}$ & 0.795$^{\pm0.002}$ & 0.704$^{\pm0.011}$ & 0.712$^{\pm0.008}$ \\
LIGHT $+$ \refineName{} & \textbf{0.765$^{\pm0.010}$} & \textbf{0.912$^{\pm0.010}$} & \textbf{0.948$^{\pm0.006}$} & 0.235$^{\pm0.076}$ & \textbf{2.642$^{\pm0.009}$} & 7.603$^{\pm0.016}$ & \textbf{0.059$^{\pm0.002}$} & \textbf{0.069$^{\pm0.003}$} & \textbf{0.181$^{\pm0.000}$} & \textbf{0.476$^{\pm0.000}$} & \textbf{0.429$^{\pm0.001}$} & \textbf{0.408$^{\pm0.002}$} & \textbf{0.808$^{\pm0.003}$} & \textbf{0.765$^{\pm0.009}$} & \textbf{0.755$^{\pm0.006}$} \\
\midrule
InterAct & 0.803$^{\pm0.010}$ & 0.928$^{\pm0.004}$ & 0.962$^{\pm0.005}$ & 0.328$^{\pm0.013}$ & 2.524$^{\pm0.037}$ & \textbf{7.827$^{\pm0.047}$} & 0.101$^{\pm0.002}$ & 0.126$^{\pm0.005}$ & 0.122$^{\pm0.001}$ & 0.418$^{\pm0.007}$ & 0.249$^{\pm0.005}$ & 0.278$^{\pm0.001}$ & \textbf{0.777$^{\pm0.003}$} & 0.631$^{\pm0.000}$ & 0.658$^{\pm0.002}$ \\
InterAct $+$ \refineName{} w/o corruption stream & 0.806$^{\pm0.005}$ & 0.929$^{\pm0.004}$ & \textbf{0.964$^{\pm0.004}$} & 0.320$^{\pm0.019}$ & 2.507$^{\pm0.039}$ & 7.836$^{\pm0.050}$ & 0.076$^{\pm0.004}$ & 0.098$^{\pm0.001}$ & 0.150$^{\pm0.001}$ & 0.421$^{\pm0.003}$ & 0.303$^{\pm0.002}$ & 0.315$^{\pm0.001}$ & \textbf{0.777$^{\pm0.001}$} & 0.670$^{\pm0.000}$ & 0.687$^{\pm0.001}$ \\
InterAct $+$ \refineName{} & \textbf{0.808$^{\pm0.006}$} & \textbf{0.930$^{\pm0.005}$} & 0.963$^{\pm0.006}$ & \textbf{0.300$^{\pm0.023}$} & \textbf{2.504$^{\pm0.040}$} & 7.842$^{\pm0.053}$ & \textbf{0.058$^{\pm0.001}$} & \textbf{0.077$^{\pm0.001}$} & \textbf{0.188$^{\pm0.000}$} & \textbf{0.430$^{\pm0.001}$} & \textbf{0.389$^{\pm0.004}$} & \textbf{0.374$^{\pm0.003}$} & \textbf{0.777$^{\pm0.004}$} & \textbf{0.749$^{\pm0.001}$} & \textbf{0.738$^{\pm0.001}$} \\
\bottomrule
\end{tabular}}
\end{center}
\end{table}

%% file: tables/refiner_supp.tex
\begin{table}[t]
\caption{\textbf{Additional metrics for the \refineName ablations in
Table~\ref{tab:ablation_refiner}} ($K{=}4$). Mean and 95\% confidence interval over three runs.}
\label{tab:refiner_supp}
\begin{center}
\resizebox{\linewidth}{!}{%
\begin{tabular}{@{}lccccc@{}}
\toprule
 & Top-2$^\uparrow$ & Top-3$^\uparrow$ & MM Dist$^\downarrow$ & Diversity$^\rightarrow$ & FSR$^\downarrow$ \\
\midrule
Ground Truth & 0.967 & 0.992 & 1.475 & 7.764 & 0.051 \\
\midrule
(a) no refinement & 0.957$^{\pm0.002}$ & \textbf{0.980$^{\pm0.002}$} & 2.198$^{\pm0.015}$ & 7.683$^{\pm0.008}$ & 0.137$^{\pm0.002}$ \\
\midrule
(b) w/o generation stream & \textbf{0.958$^{\pm0.002}$} & \textbf{0.980$^{\pm0.002}$} & 2.208$^{\pm0.015}$ & \textbf{7.696$^{\pm0.009}$} & 0.134$^{\pm0.001}$ \\
(c) w/o corruption stream & \textbf{0.958$^{\pm0.002}$} & \textbf{0.980$^{\pm0.003}$} & \textbf{2.190$^{\pm0.014}$} & 7.692$^{\pm0.008}$ & 0.085$^{\pm0.000}$ \\
(d) w/o short-range probes & \textbf{0.958$^{\pm0.002}$} & \textbf{0.980$^{\pm0.002}$} & 2.198$^{\pm0.014}$ & 7.691$^{\pm0.006}$ & 0.107$^{\pm0.001}$ \\
(e) w/o long-range probes & \textbf{0.958$^{\pm0.003}$} & \textbf{0.980$^{\pm0.003}$} & 2.202$^{\pm0.016}$ & 7.683$^{\pm0.007}$ & 0.105$^{\pm0.002}$ \\
\midrule
(f) \refineName{} & \textbf{0.958$^{\pm0.002}$} & \textbf{0.980$^{\pm0.001}$} & 2.195$^{\pm0.016}$ & 7.693$^{\pm0.007}$ & \textbf{0.078$^{\pm0.001}$} \\
\bottomrule
\end{tabular}}
\end{center}
\end{table}

%% file: tables/hparams.tex
\begin{table}[t]
\centering
\caption{Hyperparameters used by the final checkpoints.}
\label{tab:hparams}
\small
\begin{tabular}{@{}p{0.31\linewidth}p{0.65\linewidth}@{}}
\toprule
\multicolumn{2}{l}{\emph{\vaeName}} \\
Human part tokens / object tokens & 8 / 1 \\
Latent dim per token & 16 \\
Part encoder / decoder width & 1024 / 1024 \\
Spatiotemporal attention & $d=256$, 6 layers, 8 spatial / 14 temporal heads \\
Anchor set & root, left/right wrist, left/right ankle $+$ free slot \\
Base loss weights (direct recon.\ / free vel.\ / composed trans.\ / voting / human KL / object KL) & 1.0 / 1.0 / 1.0 / 0.01 / $10^{-4}$ / $10^{-4}$ \\
Auxiliary loss weights (hand / root vel.\ / foot slide / joint vel.\ / FK cons.\ / contact) & 1.0 / 50.0 / 30.0 / 30.0 / 1.0 / 1.0 \\
Training iterations & 200k; contact-tuned decoder $+$25k (encoder frozen, $\lambda_{\mathrm{ctc}}=1.0$) \\
Parameters & 755M \\
\multicolumn{2}{l}{\emph{\generatorName}} \\
Generator backbone & $d=512$, 10 layers, 8 heads, FF 2048, dropout 0.1 \\
Conditioning & text (CLIP ViT-B/32 tokens), object BPS, body-shape joint rest pose \\
Classifier-free guidance (train drop / test scale) / sampling steps & 0.1 / 2.5 / 50 \\
Optimizer & AdamW, lr 0.0001, warmup 2500, cosine to 1e-06 over 350k \\
Decoded-space losses (free pos. / free vel. / composed / alignment) & 5.0 / 5.0 / 5.0 / 2.5 \\
Training iterations & 240k \\
Parameters & 55M \\
\multicolumn{2}{l}{\emph{\refineName}} \\
Backbone & $d=256$, 4 layers \\
Short-range surface sensor & up to 64 points per query, radius 4 cm, dense surface 16,384 \\
Streams & corruption : generation = 1 : 1 (real\_every=2) \\
Recursion depth $K$ (train / test) & 3 / 4 \\
Recovery weights (absolute / centered vertices / rotations / object / root $xz$ / root $y$) & 7.0 / 5.0 / 1.0 / 1.0 / 1.0 / 1.0 \\
Generation group weights ($\lambda_c/\lambda_p/\lambda_s/\lambda_m/\lambda_n$) & 1.0 / 1.0 / 1.0 / 1.0 / 1.0 \\
Contact and penetration terms (attraction / slip / penetration) & 1.0 / 0.5 / 1.0 \\
Support terms (foot / pin and drift / leg trust) & 2.0 / 5.0 / 1.0 \\
Smoothness terms (motion / update) & 0.1 / 0.5 \\
Preservation terms (body / object / root) & 1.0 / 1.0 / 1.0 \\
Correction authority (object) & 5.0 cm, 10.0$^\circ$ \\
Optimizer & AdamW, lr 0.0003, batch 4, warmup 12k, 60k schedule (checkpoint at 44k) \\
Parameters & 4.4M \\
\bottomrule
\end{tabular}
\end{table}